\documentclass[letterpaper]{article} 
\usepackage{aaai2026}  
\usepackage{times}  
\usepackage{helvet}  
\usepackage{courier}  
\usepackage[hyphens]{url}  
\usepackage{graphicx} 
\usepackage{natbib}  
\usepackage{caption} 
\usepackage{enumitem}
\usepackage{array}
\usepackage{booktabs}
\usepackage{xcolor}
\usepackage{multirow}
\usepackage{makecell}
\usepackage{amsmath}
\usepackage{algorithm}
\usepackage{algorithmic}
\usepackage{graphicx}
\usepackage{subcaption}
\usepackage{tikz}
\usepackage{diagbox}
\newcommand{\doubleunderline}[1]{%
    \underline{\underline{#1}}%
}
\newcommand{\addArrow}[2]{%
    \begin{tikzpicture}[remember picture,baseline=-0.5ex]
        \node[inner sep=0pt] (cell) {#1}; 
        \node[anchor=south east, xshift=1pt, yshift=1pt, inner sep=0pt] 
            at (cell.south east) {#2}; 
    \end{tikzpicture}%
}
\usepackage{newfloat}
\usepackage{listings}
\DeclareCaptionStyle{ruled}{labelfont=normalfont,labelsep=colon,strut=off} 
\floatstyle{ruled}
\newfloat{listing}{tb}{lst}{}
\floatname{listing}{Listing}
\title{CoS: Towards Optimal Event Scheduling via Chain-of-Scheduling}

\author{
    Yiming Zhao,\textsuperscript{\rm 1,2} 
    Jiwei Tang,\textsuperscript{\rm 3}
    Shimin Di,\textsuperscript{\rm 4,5}
    Libin Zheng,\textsuperscript{\rm 1}\thanks{Corresponding author: zhenglb6@mail.sysu.edu.cn}
    Jianxing Yu,\textsuperscript{\rm 1}
    Jian Yin\textsuperscript{\rm 1}
}

\affiliations{
\textsuperscript{\rm 1}School of Artificial Intelligence, Sun Yat-sen University\\
\textsuperscript{\rm 2}Key Laboratory of Intelligent Assessment Technology for Sustainable Tourism, Ministry of Culture and Tourism, Sun Yat-sen University\\
\textsuperscript{\rm 3}Shenzhen International Graduate School, Tsinghua University\\
\textsuperscript{\rm 4}School of Computer Science and Engineering, Southeast University\\
\textsuperscript{\rm 5}Key Laboratory of New Generation Artificial Intelligence Technology and Its Interdisciplinary Applications\\
}

\usepackage{bibentry}

\begin{document}

\maketitle
\begin{abstract}
Recommending event schedules is a key issue in Event-based Social Networks (EBSNs) in order to maintain user activity. An effective recommendation is required to maximize the user's preference, subjecting to  both time and geographical constraints. Existing methods face an inherent trade-off among efficiency, effectiveness, and generalization, due to the NP-hard nature of the problem.
This paper proposes the \textbf{Chain-of-Scheduling (CoS)} framework, which activates the event scheduling capability of Large Language Models (LLMs) through a guided, efficient scheduling process. CoS enhances LLM by formulating the schedule task into three atomic stages, \textit{i.e.}, \textit{exploration}, \textit{verification} and \textit{integration}. Then we enable the LLMs to generate CoS autonomously via Knowledge Distillation (KD). Experimental results show that CoS achieves near-theoretical optimal effectiveness with high efficiency on three real-world datasets in a interpretable manner. Moreover, it demonstrates strong zero-shot learning ability on out-of-domain data. 
\end{abstract}
\begin{links}
    \link{Code}{https://github.com/kiki123-hi/CoS}
\end{links}
\section{Introduction}
\begin{figure}[h!]
    \centering
    \begin{subfigure}[b]{1\columnwidth}
        \centering
        \includegraphics[width=\textwidth]{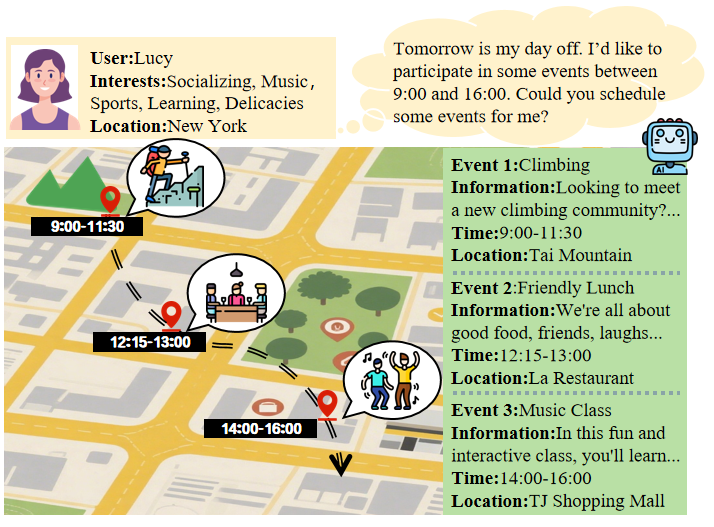}
        \caption{A specific example of Event Scheduling.}
        \label{fig:intro}
    \end{subfigure}
    \par
    \begin{subfigure}[b]{1\columnwidth}
    \centering
    \includegraphics[width=\textwidth]{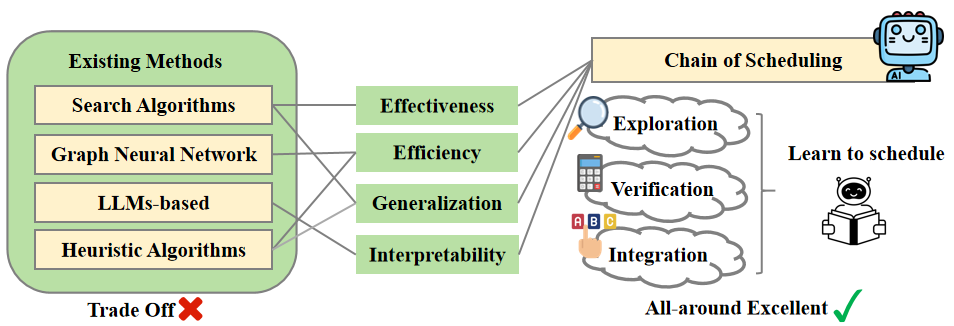}
        \caption{Challenges of Event Scheduling.}
        \label{fig:challenge}
    \end{subfigure}
    \caption{A illustration of event scheduling. (a) is a specific example of event scheduling. (b) denotes the challenges of event scheduling, \textit{i.e.}, existing methods face a trade-off among effectiveness, efficiency, and interpretability while CoS can achieve all-round excellent via \textit{exploration}, \textit{verification}, and \textit{integration}.}
    \label{fig:event_scheduling_task_resized}
\end{figure}


As an emerging online social network, event-based social networks (EBSNs) enable users to organize and participate in offline social activities. Platforms such as Meetup, Eventbrite, and Douban City have more than 10 million monthly active users, playing a significant role in bridging the gap between online users to meet and communicate in the physical world. To maintain user activity, the core task behind these platforms is to effectively recommend events to users, which typically advises groups of events per user based on their interests. The recommended event group is usually organized as a schedule, allowing the user to participate offline in them from one to another (Figure~\ref{fig:event_scheduling_task_resized}(a) gives an example). Statistics show that Meetup has more than 16 million users who participate in over 300,000 events each month, thanks to its personalized event recommendation system~\cite{DBLP:journals/tkde/ChengYCGWL21}. Thus, effectively scheduling events for users is a key issue for EBSNs.

Current methods for event scheduling problem can be categorized into three main types based on their primary concerns: effectiveness, efficiency, and generalization.

\begin{figure*}[htb] 
  \centering 
  \includegraphics[width=1\textwidth]{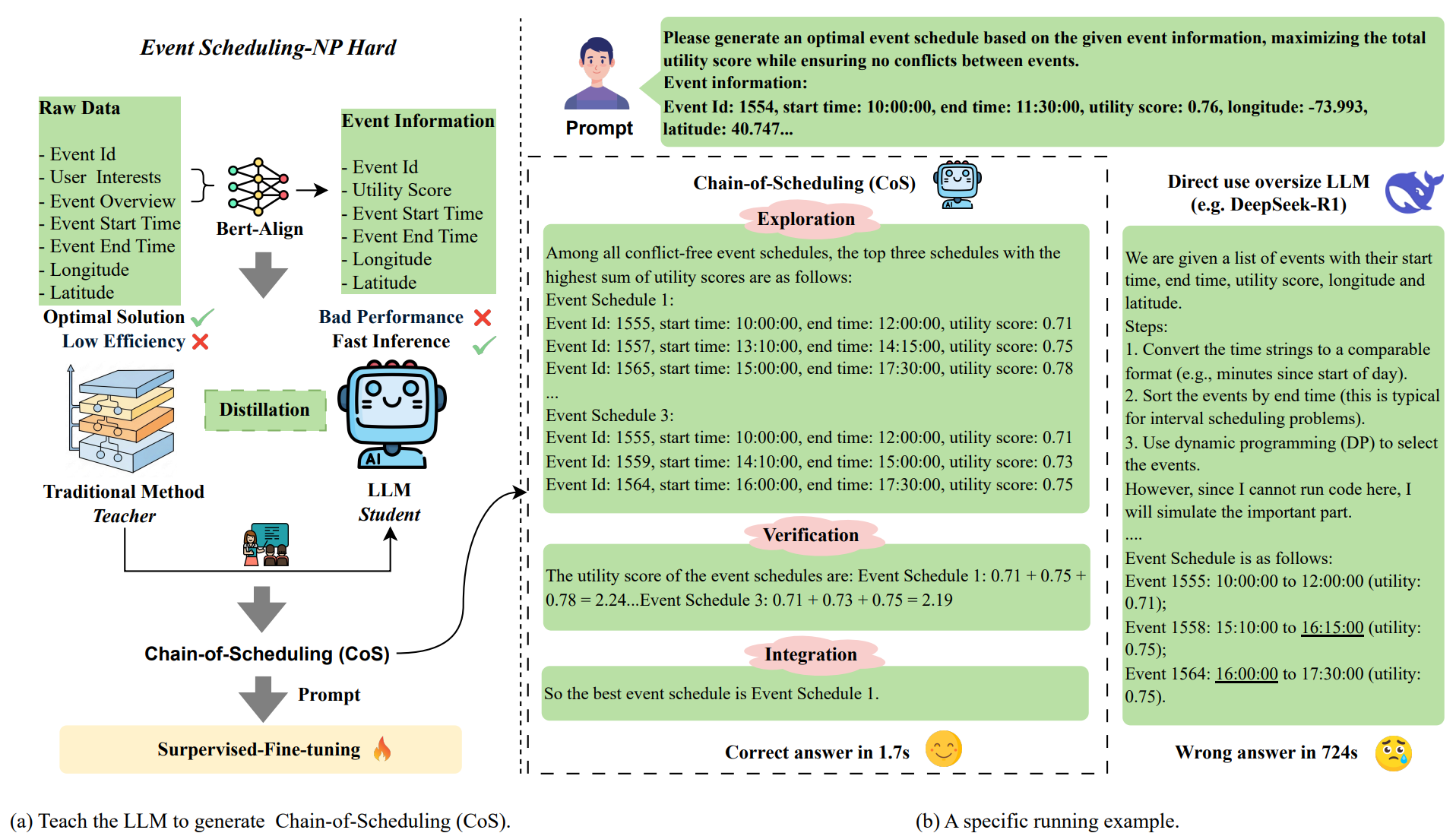} 
  \caption{\textbf{The illustration of the Chain-of-Scheduling (CoS).} (a) denotes the basic process of teaching the LLM to generate CoS. Distillation CoS from Combinatorial Optimization, then supervised fine-tune (SFT) the (prompt, CoS) pairs, enabling the LLM to autonomously generate CoS. (b) is a specific running example. Small LLM (\textit{e.g.}, Qwen2.5-7B-Instruct) can quickly arrive at the correct answer via CoS, while directly using a oversized LLM (\textit{e.g.}, DeepSeek-R1) will fall into an overly lengthy and time-consuming thinking process, ultimately leading to an incorrect answer.}
  \label{fig:framework} 
\end{figure*}

\textbf{Effectiveness-targeted methods} aim to achieve higher schedule quality by precisely matching users' preferences subjecting to the complex constraints. Among these, search algorithms like grid search and dynamic programming~\cite{DBLP:journals/ijsysc/ZhangZLDH23}  find optimal solutions but face at least quadratic time complexities, leading to its inefficiency for large problem inputs. In recent years, Large Language Models (LLMs) have emerged as an area of research in the planning domain, with prompt engineering techniques like Chain-of-Thought (CoT) achieving significant progress in various Natural Language Processing (NLP) tasks~\cite{DBLP:journals/corr/abs-2412-15115,DBLP:journals/corr/abs-2501-12948}. Nevertheless, when it comes to event scheduling, even with long CoTs often suffer from redundancy, lack of focus, and overthinking~\cite{DBLP:journals/corr/abs-2412-21187,DBLP:journals/corr/abs-2503-16419,DBLP:journals/corr/abs-2505-00127}, leading to time-consuming exploration and invalid schedules (Figure~\ref{fig:framework}(b) gives an example). Automated agent workflow generation methods~\cite{DBLP:conf/iclr/ZhangXYTCCZCHWZ25,DBLP:conf/aaai/HuangLC25,DBLP:journals/corr/abs-2509-05941,DBLP:journals/corr/abs-2509-21834} struggle to fully understand the complex constraints, making their planning behaviors abnormal or fall into ineffective loops~\cite{DBLP:journals/corr/abs-2409-13373,DBLP:journals/corr/abs-2403-04121,DBLP:journals/corr/abs-2502-12435}.

\textbf{Efficiency-targeted methods} focus on efficiently generating schedules, typically employing heuristic algorithms to hold a low computational complexity, such as greedy algorithms~\cite{DBLP:journals/tkde/ChengYCGWL21} and genetic algorithms~\cite{DBLP:journals/evi/AlhijawiA24}. However, these methods often sacrifice effectiveness, and face lower schedule quality regarding users' preferences. Graph Neural Networks (GNNs) \cite{DBLP:journals/tsmc/LiuH23,DBLP:conf/aaai/ChenTT24} are constrained by their small parameter scale and black-box nature, making it difficult to model complex event relationships. In addition, their lack of interpretability is also a serious concern for usage.

In terms of \textbf{generalization}, aforementioned methods are generally stable and yield consistent behaviors on unseen data, but limited to either effectiveness or efficiency. In contrast, GNNs struggle to learn the underlying logic and general rules behind the data, preventing them from effectively transferring existing knowledge and experience when faced with unseen data distributions, leading to poor generalization~\cite{DBLP:conf/nips/MaDM21,DBLP:conf/www/WuNYBY24}.

Event scheduling for EBSN, which requires considering both temporal and spatial constraints, is an NP-hard problem (we prove in Appendix B). As illustrated in Figure~\ref{fig:event_scheduling_task_resized}(b), the aforementioned methods inevitably face a hard trade off on certain aspects. This naturally leads to a research question: \textit{How can we achieve effective and efficient event scheduling in an interpretable manner while maintaining strong zero-shot learning capabilities?}

To this end, we propose Chain-of-Scheduling (CoS), which specifically standardizes the scheduling task into three atomic stages, \textit{exploration}, \textit{verification}, and \textit{integration}, while fully leveraging the interpretability of LLMs. Different from CoT, CoS is not a general-purpose thinking process. Instead, it provides a preset, rigorously structured guidance framework. 
It significantly boosts computational efficiency by formulating efficient reasoning paths for LLMs and minimizing unproductive inference steps, thus avoiding the degradation of CoT performance caused by redundant thinking and lack of clear direction.
Specifically, the \textit{exploration} step guides LLMs in targeted exploration, drastically narrowing the ineffective search space and enabling LLMs to more \textit{efficiently} find high-quality candidate solutions. The \textit{verification} and \textit{integration} steps then guide LLMs to evaluate candidate solutions and make optimal choices, ensuring the \textit{effectiveness} of the generated schedule while forming a complete, coherent, and \textit{interpretable} scheduling solution. To enable LLMs to learn to schedule, we perform Knowledge Distillation (KD)~\cite{DBLP:journals/ijcv/GouYMT21}. Specifically, we use search algorithms as teacher models, constructing the high quality solutions into a natural language format. Then we distill this scheduling knowledge into LLMs via Supervised Fine-tuning (SFT). Through this process, LLMs internalize rich spatiotemporal knowledge and complex constraint handling logic, allowing them to \textit{generalize} to unseen datasets. 

Our contributions are three-fold:
\begin{itemize}
    \item 
    We propose CoS, a framework that integrates \textit{exploration}, \textit{verification} and \textit{integration} to achieve all-round excellence in event scheduling across effectiveness, efficiency, generalization, and interpretability.
    \item Through Knowledge Distillation (KD), We enable the LLMs to generate CoS autonomously, which guides LLMs to internalize spatiotemporal knowledge and complex constraint handling logic for event scheduling.
    \item Experimental results show our method significantly outperforms other existing methods in terms of efficiency and effectiveness, and demonstrates strong generalization ability on out-of-domain unseen datasets.
    
\end{itemize}
\section{Preliminaries}
\paragraph{Event Scheduling Data.} An event refers to an offline activity that a user may participate in, with the  $i$-th event  denoted as $e_i$. All the announced events on the EBSN comprise a set as $E=\{e_i\}$. Each event $e_i = \langle loc_i, t_i^{start}, t_i^{end}  \rangle$ is associated with a geographical location $loc_i $ (where the event is held), and the starting \& ending time $t_i^{start} $, $t_i^{end}$.   Users give varying preferences over the events. For each user $u_j$, there is a utility score $s_{i,j}$  representing their preference for  $e_i$.  In practice, $s_{i,j}$ is typically computed according to the profiles of events and users, and each EBSN develops its own model for $s_{i,j}$ regarding the platform characteristics~\cite{DBLP:conf/kdd/LiLBLY14,DBLP:conf/sigmod/SheT015,DBLP:journals/tkde/SheTCC16}. Thus, following existing works~\cite{DBLP:conf/kdd/LiLBLY14,DBLP:conf/sigmod/SheT015,DBLP:journals/tkde/SheTCC16,DBLP:conf/icde/ChengYCGW17,DBLP:journals/tkde/ChengYCGWL21} , we focus on how to effectively recommending and scheduling the events, and treat $s_{i,j}$  as our input. Note that our developed model is orthogonal to the computation of $s_{i,j}$'s, and is applicable for all EBSNs.

\paragraph{Event Scheduling.} Given an event set $E$ and a user $u_j$, the recommendation task on EBSN is to generate an event sequence $T^* = \langle e_{i_1} \rightarrow  e_{i_2}  \rightarrow \cdots \rangle$ that maximizes the total utility score while retaining validity.

We first define the \emph{feasible solution space} as:
\begin{gather}
\mathcal{T} \;=\;
\Bigl\{\, 
T = \langle e_{i_1}\!\rightarrow e_{i_2}\!\rightarrow \dots\rangle 
\;\Bigr\} \, .\\
    \text{s.t. } 
\forall\, (e_{i} \rightarrow  e_{i'}) \in T\,,\quad  t_{i'}^{start} - t_{i}^{end} \ge  t(loc_{i}, loc_{i'}) \, , \nonumber 
\end{gather}
where the function $t(\cdot)$ denotes the traveling time between the two locations. 

Then, the optimal event sequence $T^*$ is given by

\begin{gather}
      T^* = \arg\max_{T \in \mathcal{T}}  \sum\limits_{e_i \in T} s_{i,j} \, . 
\end{gather}

\section{Related Work}

\paragraph{Large Language Models.} The emergence of large language models has transformed the research paradigm in the Natural Language Processing (NLP) community~\cite{DBLP:conf/naacl/TangXLZZHZ25,DBLP:journals/corr/abs-2505-12215,DBLP:conf/acl/Guo0DSC025}, especially the oversized large models such as DeepSeek and Qwen~\cite{DBLP:journals/corr/abs-2501-12948,DBLP:journals/corr/abs-2412-19437,DBLP:journals/corr/abs-2412-15115}. Recently, the concept of test-time computation has gained popularity, with many studies employing chain of thought and reinforcement learning on top of base models to further scale up performance at test time, achieving impressive results across various tasks. However, when directly applied to planning tasks, these Oversized large models often perform poorly, even failing to complete simple actions~\cite{DBLP:journals/corr/abs-2409-13373,DBLP:journals/corr/abs-2403-04121,DBLP:journals/corr/abs-2502-12435}. In our event scheduling tests, Oversized large models often take more than ten minutes to process a planning scenario with fewer than 20 total events, only to produce incorrect results. \textit{Therefore, the planning potential of LLMs remains largely untapped.}

\paragraph{Combinatorial Optimization.} Although event scheduling is a NP-hard problem, there are still some Combinatorial Optimization that can find exact solutions, such as grid search and dynamic programming~\cite{DBLP:journals/eor/BastosMHF19}. However, these methods have high time complexity, above quadratic level, and are unable to handle a large number of event inputs. There are also some heuristic algorithms, such as the greedy algorithm~\cite{DBLP:journals/tkde/ChengYCGWL21} that searches for local optima and can solve event scheduling relatively quickly, and the genetic algorithm~\cite{DBLP:journals/evi/AlhijawiA24} that searches for global optima but at a slower pace. Yet, the results of these methods often have a significant gap from the optimal solution. \textit{Therefore, Combinatorial Optimization either face a trade-off between time efficiency and effectiveness or fail to achieve satisfactory performance in both aspects, making it difficult to realize optimal event scheduling.}

\paragraph{Traditional Deep Learning-based Methods.} Event scheduling can be formalized as a weighted activity
selection problem and modeled by GNNs. Prior works include heuristic graph search algorithms that learn domain-independent heuristic functions via GNNs~\cite{DBLP:conf/aaai/ChenTT24}, and reinforcement learning-based GNNs that can model event scheduling as Markov decision processes to learn state representations and train agents for near-optimal decisions~\cite{DBLP:journals/tsmc/LiuH23}. \textit{However, these methods are limited by the relatively small parameter scale and modeling capacity of GNNs, leading to poor performance in event scheduling problems. Moreover, they lack interpretability, essentially operating as black boxes to select the optimal set of events that satisfy the constraints.}

\section{Method}
\label{sec:method}
As shown in Figure \ref{fig:framework} (b), raw LLMs are still weak in terms of generating high-quality event schedules. To upgrade LLMs for such a task, as shown in Figure \ref{fig:framework} (a), we devise the Chain-of-Scheduling (CoS) framework, which treats it as a student model and teaches it to schedule with a chain of reasoning. In particular, CoS consists of two components. The first is CoS Construction component, preparing a set of chain-of-scheduling data (the reasoning and evidence for generating high-quality schedules),  which would be used to teach/train the LLM in the second component. The second component is the CoS Knowledge Distillation component, which distills the constructed schedule knowledge chain into the LLM. 
\subsection{The Construction of CoS}
\label{subsec:construct_cos}

We first describe how we construct the CoS data to be used in the Knowledge Distillation stage.

Chain-of-Scheduling (CoS) formalizes the schedule reasoning process as a composition of three atomic stages: 1) \textit{Exploration}: exploring high-quality schedules, 2) \textit{Verification}: verifying the utility score of each solution, and 3) \textit{Integration}: integrate the current evidence and select the best solution.

\textit{Exploration.} To mimic the human's reasoning for generating a satisfying schedule, this step first enumerates $k$ high-quality schedules, which are valid in the first place. They could be either the exact top-$k$ schedules or the approximated top-$k$ ones because of the time complexity variance. In terms of the former, they are denoted as
\begin{equation}
    T_{\text{top-}k} = \arg{\mathrm{top}\text{-}k}_{T \in \mathcal{T}} \left( \sum_{e_{i} \in T} s_{i,j} \right) \, ,
\end{equation}
where $T_{\text{top-}k}$ is top-$k$ candidate schedules.

In the offline training step, we employ grid search or dynamic programming (DP)~\cite{DBLP:journals/ijsysc/ZhangZLDH23} to collect such top-$k$ schedules for each training input instance. In this way, the effectiveness-targeted algorithms like DP serve as the teacher model while the LLM serves as the student model. The insight lies in: although search algorithms like dynamic programming may be computationally expensive, they reliably yield optimal or near-optimal solutions. This motivates us to take it as a knowledge source for teaching LLMs.

\textit{Verification.} After exploration, we mimic a human's verification thinking by evaluating each solution's quality, \textit{i.e.}, 
explicitly computing the utility score for each schedule. For example, if a schedule contains two events with utility scores 2 and 3, the verification step generates the reasoning trace: ``2 + 3 = 5". This can be formalized as

\begin{equation}
    v(T) = \sum_{e_i \in T} s_{i,j} \, ,
\end{equation}
where $v(S)$ is the verified total utility score of schedule $s_{i,j}$.

\textit{Integration}. Once exploration and verification are complete, we have multiple near-optimal schedules along with their verified utility scores. The final step is to pick the maximum:
\begin{equation}
T^* = \text{argmax}_{T \in T_{\text{top-}k}} v(T) \, ,
\end{equation}
where $T^*$ is the optimal event schedule with maximum utility score.

\subsection{CoS Knowledge Distillation}
\label{subsec:cos_kd}
The core goal of this phase is to distill the structured reasoning capability of the CoS framework into LLMs. This is achieved by training LLMs to autonomously generate CoS traces given input event sets and user contexts, framed as knowledge distillation: meticulously constructed CoS traces (from algorithms like dynamic programming) serve as the teacher model, while LLMs act as student models mimicking the teacher's step-by-step reasoning. Through SFT, this high-quality reasoning knowledge is transferred to LLMs. Leveraging LLMs' inherent efficient reasoning (e.g., parallel decoding) and inference acceleration frameworks (e.g., vLLM), SFT-tuned LLMs can reproduce CoS reasoning with ultra-low latency, enabling them to inherit traditional methods' high-quality planning capabilities while addressing computational inefficiency.

\paragraph{CoS Alignment.}
We construct a specialized SFT dataset $D_{\text{SFT}} = \{(x_i, y_i^{\text{CoS}})\}_{i=1}^N$ to bridge CoS reasoning and LLM text generation. Each input $x_i$ is a complete event scheduling problem instance, containing an event set, user context, and preference information. The target output $y_i^{\text{CoS}}$ is the complete CoS reasoning trace generated for $x_i$ by search algorithm-based CoS construction (as shown in Figure 2 (b)). Natural language rationales in $y_i^{\text{CoS}}$ bridge the gap between symbolic-logical constructs and LLM-compatible textual reasoning traces.

\paragraph{SFT Objective Function.}
The goal of SFT is to maximize the likelihood of LLMs generating correct CoS traces, achieved by minimizing token-level cross-entropy loss over the entire CoS. For a single instance $(x_i, y_i^{\text{CoS}})$, the loss is:
\begin{equation}
    \mathcal{L}_{\text{SFT}}^{(i)}(\theta) = -\sum_{t=1}^{L_i} \log P_{\text{LLM}}(y_{i,t}^{\text{CoS}} \mid x_i, y_{i,<t}^{\text{CoS}}; \theta) \, ,
\end{equation}
where $L_i$ is the token length of $y_i^{\text{CoS}}$, and $y_{i,t}^{\text{CoS}}, y_{i,<t}^{\text{CoS}}$ denote the $t$-th token and preceding tokens of $y_i^{\text{CoS}}$, respectively. The overall SFT objective across the dataset is:
\begin{equation}
    \mathcal{L}_{\text{SFT}}(\theta) = \mathbf{E}_{(x, y^{\text{CoS}}) \sim \mathcal{D}_{\text{SFT}}} \left[ \mathcal{L}_{\text{SFT}}^{(i)}(\theta) \right] \, .
\end{equation}
Minimizing $\mathcal{L}_{\text{SFT}}(\theta)$ trains LLMs to accurately predict CoS sequence tokens, replicating structured reasoning and distilling traditional methods' planning capabilities. 

\subsection{Schedule Post-processing}
\begin{figure}[t] 
    \centering 
    \includegraphics[width=1\columnwidth]{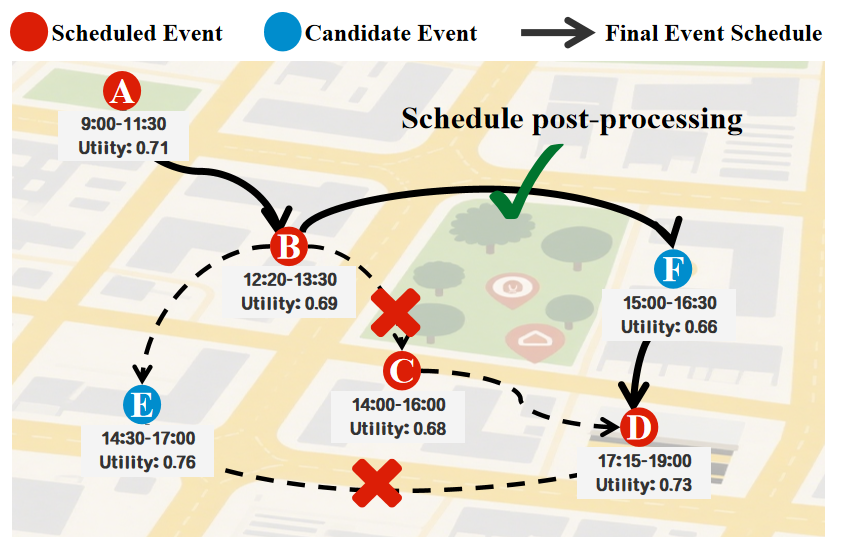} 
    \caption{A post-processing example.} 
    \label{fig:post} 
\end{figure}
As above, we distill the spatiotemporal knowledge essential for event scheduling to LLM, to achieve both effectiveness and efficiency. However, since the hallucination issue could not yet be eliminated for the state-of-the-art LLMs~\cite{DBLP:journals/csur/JiLFYSXIBMF23,DBLP:journals/csur/ChakrabortyOD25}, the fine-tuned LLM with CoS is also likely to produce schedules that are not all valid. As later shown by the experiment results, though LLM equipped CoS already achieves a large improvement in terms of the event conflicts (two events are not compatible in terms of time or distance), a valid schedule is not always guaranteed. We thereby propose a light-weight post-processing step, which further confirms the  validity of the schedule. 

As shown in Figure~\ref{fig:post}, the original generated schedule is $A\rightarrow B\rightarrow C \rightarrow D$, while event $B$ and $C$ are too close in time while far away in terms of distance, preventing the user from reaching $C$ in time after finishing $B$. To eliminate this conflict, our post-processing would perform a local search which anchors on the first conflicted event to find substitution for the successor event. In the example, it finds the event $F$ which is compatible with both $B$ and $D$, leading to the new schedule $A\rightarrow B\rightarrow F \rightarrow D$ with a slight utility loss.

\begin{table*}[tb]
    \centering
    \small
    \setlength{\tabcolsep}{1.5mm}
    \begin{tabular}{l|ccc|ccc|ccc}
    \toprule
        \multirow{2}{*}{\textbf{Methods}} & \multicolumn{3}{c}{\textbf{NewYork}} & \multicolumn{3}{c}{\textbf{Washington}} & \multicolumn{3}{c}{\textbf{London}} \\
        \cmidrule{2-10}
        & \multicolumn{1}{c}{\addArrow{\shortstack{Utility\\ Score}}} 
        & \multicolumn{1}{c}{\addArrow{\shortstack{Latency\\(seconds)}}} 
        & \multicolumn{1}{c|}{\addArrow{\shortstack{Conflict\\Rate (\%)}}} 
        & \multicolumn{1}{c}{\addArrow{\shortstack{Utility\\ Score}}} 
        & \multicolumn{1}{c}{\addArrow{\shortstack{Latency\\(seconds)}}} 
        & \multicolumn{1}{c|}{\addArrow{\shortstack{Conflict\\Rate (\%)}}} 
        & \multicolumn{1}{c}{\addArrow{\shortstack{Utility\\ Score}}} 
        & \multicolumn{1}{c}{\addArrow{\shortstack{Latency\\(seconds)}}} 
        & \multicolumn{1}{c|}{\addArrow{\shortstack{Conflict\\Rate (\%)}}}  \\
    \midrule

    \multicolumn{10}{@{}l}{\textbf{\textit{Standard pretrained LLMs}}} \\
    DeepSeek-R1 & 2.97 & 785 & \doubleunderline{66} & 2.99 & 496 & 98 & 3.01 & 750 & 99 \\
    DeepSeek-V3 & 2.45 & 48.7 & 90 & 2.71 & 46.5 & 89 & 2.68 & 35.4 & 94 \\
    Qwen-Max & 2.44 & 141 & 97 & 2.63 & 197 & 98 & 2.81 & 145 & 97 \\
    Qwen-Plus & 2.23 & 34.3 & 98 & 2.43 & 38.1 & 99 & 2.11 & 42.7 & 99 \\

    \cmidrule(r){1-1}\cmidrule(lr){2-10}
    
    \multicolumn{10}{@{}l}{\textbf{\textit{Combinatorial Optimization}}} \\
    Grid Search & 3.73 & $>10^3$ & -- & 4.02 & $>10^3$ & -- & 5.17 & $>10^3$ & -- \\
    Dynamic Programming & 3.73 & 14.2 & -- & 4.02 & 8.58 & -- & 5.17 & 24.6 & -- \\
    Greedy & 1.95 & \textbf{0.01} & -- & 2.31 & \textbf{0.01} & -- & 2.51 & \textbf{0.01} & -- \\
    Genetic Algorithm & 1.14 & 3.79 & -- & 1.07 & 4.14 & -- & 1.12 & 9.86 & -- \\

    \cmidrule(r){1-1}\cmidrule(lr){2-10}
    
    \multicolumn{10}{@{}l}{\textbf{\textit{Deep learning-based Methods}}} \\
    GOOSE & 3.15 & 59.2 & -- & 3.02 & 58.8 & -- & 2.93 & 60.5 & -- \\
    GNN-DRL & \doubleunderline{2.85} & \underline{0.44} & -- & \doubleunderline{2.64} & \underline{0.45} & -- & \doubleunderline{3.18} & \doubleunderline{4.23} & -- \\

    \cmidrule(r){1-1}\cmidrule(lr){2-10}
    
    \multicolumn{10}{@{}l}{\textbf{\textit{LLMs with multi-step reasoning}}} \\
    DeepSeek-V3 + Chain-of-Thought & 2.48 & 172 & 86 & 2.58 & 104 & \doubleunderline{84} & 2.54 & 108 & \doubleunderline{88} \\
    Qwen-Plus + Chain-of-Thought & 2.16 & 30.1 & 94 & 2.44 & 38.1 & 98 & 2.19 & 50.6 & 96 \\ 
    Qwen2.5-7B + Chain-of-Thought & 2.14 & 2.32 & 99 & 2.32 & 4.72 & 99 & 2.41 & 7.54 & 99 \\
    Mistral-7B + Chain-of-Thought & 2.12 & 2.27 & 98 & 2.19 & 2.57 & 97 & 2.22 & 5.07 & 96 \\
    Qwen2.5-7B + Chain-of-\textbf{Scheduling} & \underline{3.37} & \doubleunderline{1.29} & \underline{15} & \underline{3.57} & \doubleunderline{1.39} & \underline{22} & \underline{4.50} & \underline{3.36} & \underline{41} \\
    Mistral-7B + Chain-of-\textbf{Scheduling} & \textbf{3.44} & 1.84 & \textbf{9.8} & \textbf{3.64} & 1.94 & \textbf{15} & \textbf{4.65} & 4.79 & \textbf{19} \\
    \bottomrule
    \end{tabular}
    \caption{Performance comparison of different methods. We \textbf{bold} the best, \underline{underline} the second best and \doubleunderline{double underline} the third among the efficiency-aware methods (i.e., except the time-costly exact solutions)}
    \label{tab:main_result}
\end{table*}

\begin{table}[tb]
    \centering
    \small
    \setlength{\tabcolsep}{2mm}
    \begin{tabular}{l|c|c}
    \toprule
        \diagbox[width=16em, height=2\normalbaselineskip]{\textbf{Method}}{\textbf{Utility Score}} 
        & \textbf{Washington} 
        & \textbf{London} \\
    \midrule
        \multicolumn{1}{@{}l}{\textbf{\textit{Deep learning-based Methods}}} & & \\
        GOOSE & 3.32 & 2.85 \\
        GNN-DRL & 2.60 & 3.04 \\
        \cmidrule(r){1-1}\cmidrule(lr){2-3}
        \multicolumn{1}{@{}l}{\textbf{\textit{LLMs with multi-step reasoning}}} & & \\
        Qwen2.5-7B + Chain-of-\textbf{Scheduling} & 3.65 & 4.22 \\
        Mistral-7B + Chain-of-\textbf{Scheduling} & 3.47 & 3.87 \\
    \bottomrule
    \end{tabular}
    \caption{Generalization test results}
    \label{tab:generalization}
\end{table}
\section{Experiments}
\label{sec:exp}

In this section, we attempt to answer the following four research questions (RQs): 
\begin{itemize}
    \item \textbf{RQ1:} How does Chain-of-Scheduling (CoS) perform compared to baselines?
    \item \textbf{RQ2:} How effective is the zero-shot learning capability of CoS?
    \item \textbf{RQ3:} How effective are the various components within CoS?
    \item \textbf{RQ4:} How does the parameter $k$ in the \textit{Exploration} step affect CoS?
\end{itemize}

\subsection{Experimental Setting}

\paragraph{Dataset Description.}
To evaluate the effectiveness of the proposed model across different urban computing scenarios, we crawl data from the Meetup website from 2-year period on three cities, \textit{i.e.}, New York, Washington, and London.
We eventually obtain 74411, 81395, 218773 events and 45854, 44742, and 22381 users for the cities respectively. In this way, we evaluate the methods across different urban scenarios, investigating their robustness. For each city, the data is further divided into training and test sets in a 4:1 ratio. To form a training/testing sample, we randomly draw a user and a date, and then obtain the corresponding events on the date. An input instance is then formulated as recommending an event schedule for the user on the date (from 9 am to 9 pm). In particular, for testing, we repeat more than 1600 evaluations and report the average performance. 

An event includes three features: event ID, the time window, and the event description. A user contains features: user ID, interests, and the user's attended events. For each evaluation input, the locations and time features for the events could be directly obtained from the dataset. Since the platforms do not directly disclose users' preferences over events, we simulate the utility score between an event and a user. We use BERT~\cite{DBLP:conf/naacl/DevlinCLT19} to align the event/user descriptions and then compute their semantic similarity, obtaining a utility score ranging from 0 to 1. Note that our method is orthogonal to any utility formulation, and our simulation in this experiment effectively captures users' ground-truth participation over events (see demo in Appendix A).

\paragraph{Base Models.}
To fully the validate effectiveness and superiority of our method, we conduct experiments with two types of base models: one is the dense model Qwen2.5-7B-Instruct~\cite{DBLP:journals/corr/abs-2412-15115}, and the other is the sparse model Mistral-7B-Instruct-v0.3~\cite{rubra_ai_2024} based on mixed-of-experts (MoE).

\paragraph{Hyperparameters Settings.}
In the construction phase of Chain-of-Scheduling (CoS), we select the top $k$ event schedules with the highest utility scores, where $k$ is set to 3 by default.

Our method employs lightweight Low-Rank Adaptation, \textit{i.e.}, LoRA training approach~\cite{DBLP:conf/iclr/HuSWALWWC22}, with specific parameter settings as follows: the maximum length of the model is set to 32,768, the learning rate is set to 1e-5, the number of epochs is set to 3, the batch size is set to 2, the alpha value of lora is set to 16, and the rank value is set to 8. The entire training process is carried out on two NVIDIA A800-SXM4-80GB GPUs.

\paragraph{Baselines.}  
We compare with three categories of methods: Pre-trained LLMs (see Appendix C.1 for the prompts), combinatorial optimization, and traditional deep learning-based methods.  We provide a description for each baseline below.

\textbf{Standard pretrained LLMs:} 
\begin{itemize}
    \item \underline{Qwen2.5-Plus}~\cite{DBLP:journals/corr/abs-2412-15115}: an optimized version of Qwen2.5 by Alibaba, with strong language capabilities, outperforming DeepSeek-V2.5.
    \item \underline{Qwen2.5-Max}~\cite{DBLP:journals/corr/abs-2412-15115}: an MoE model by Alibaba with over 100 billion parameters and 100k context length.
    \item \underline{DeepSeek-V3}~\cite{DBLP:journals/corr/abs-2412-19437}: an open-source model by DeepSeek, with 671 billion parameters and 128k token context window.
    \item \underline{DeepSeek-R1}~\cite{DBLP:journals/corr/abs-2501-12948}: a reasoning model by DeepSeek, trained via reinforcement learning.
\end{itemize}
For LLM-based methods, since 
their generated schedule may be invalid, 
we adopt the same post-processing proposed in Section 4.3 to ensure fairness.

\textbf{Combinatorial Optimization:} 
\begin{itemize}
    \item \underline{Greedy}~\cite{DBLP:conf/kdd/LiLBLY14,DBLP:journals/tkde/SheTCC16,DBLP:conf/icde/ChengYCGW17,DBLP:journals/tkde/ChengYCGWL21}: it selects events with the highest utility score without conflicts.
    \item \underline{Grid Search:} it enumerates all solutions to find the optimal conflict-free task set.
    \item \underline{Dynamic Programming}~\cite{DBLP:journals/ijsysc/ZhangZLDH23}: it breaks down complex problems to maximize total utility score.
    \item \underline{Genetic Algorithm}~\cite{DBLP:journals/evi/AlhijawiA24}: it optimizes event sequences using natural selection.
\end{itemize}

\textbf{Traditional deep learning-based methods:} 
\begin{itemize}
    \item \underline{GOOSE}~\cite{DBLP:conf/aaai/ChenTT24}: it uses GNNs to guide classical planning search.
    \item \underline{GNN-DRL}~\cite{DBLP:journals/tsmc/LiuH23}: it solves DJSSP using GNNs and DRL, minimizing total processing time.
\end{itemize}

\paragraph{Evaluation Metrics.} We comprehensively evaluate the performance of various methods in handling event scheduling from three perspectives. \textbf{Utility Score} represents the user's average preference for each given sequence of events, with higher values indicating better. 
\textbf{Latency} evaluates the running time of each method. 
\textbf{Conflict Rate} measures the proportion of events conflicting with each other in a schedule. Remember that to ensure a valid schedule, for two adjacent events in the sequence, their time windows should not overlap and retain a gap allowing the user to physically travel from one to another (otherwise resulting in a conflict). In our evaluation, we find that LLM-based methods cannot ensure 0\% conflicts, so we have a post-processing step as mentioned before. Nevertheless, we still report their conflict rates before the post-processing step, serving as a complementary metric to explain the methods' performances.

\subsection{Main Results (RQ1)}
In this section, we report the comparison results across three different cities: New York, Washington, and London. 
We can draw four key findings:
\begin{itemize}
    \item \textbf{Standard pretrained LLMs generally suffer low utility scores  and high latency.} The reason for the low utility scores could be reflected by their conflict rates, which are dramatically larger than other methods. Even with chain-of-thoughts, they are still struggling for generating schedules where events are compatible with each other. 
    They typically have higher latency due to API calls.
    \item \textbf{Combinatorial Optimization fail to achieve both satisfying latency and utility.} Although grid search and dynamic programming can obtain optimal solutions, their time complexity is above the quadratic level, leading to long delays. The greedy algorithm is quick but has a utility score far from the optimum. Genetic algorithm, on the other hand, is weak in either aspects.
    \item \textbf{Graph Neural Networks-based methods are not satisfactory either.} GNN-DRL is fast but achieve inferior utilities, achieving  $61\sim76\%$ of the optimum. GOOSE is better on New York and Washington, but suffers high latency. In particular, the ratio between their utility and the optimum decay over the three datasets, from  New York to London. This indicates their performances become worse on larger inputs.
    \item \textbf{CoS maintains low planning latency while significantly surpassing other approximation methods in  utility across all datasets.} The utility of CoS on all datasets remain above $90\%$ of the optimum, far higher than other approximation methods. The planning latency is kept within 2 seconds on datasets with fewer events, i.e., New York and Washington, and within 5 seconds on London. The ourstanding peformance of CoS is attributed to its low conflict rate compared to others.
\end{itemize}  
Some model responses are provided in Appendix C.2.

\subsection{Zero-Shot Event Scheduling Performance (RQ2)}

In this section, we examine the predictive performance of the proposed model in zero-shot scenarios, with the results presented in Table~\ref{tab:generalization}. Specifically, we train the methods on the New York dataset and test them on the Washington and London datasets.

The results in Table 2 highlight CoS's exceptional performance in event scheduling on unseen cities London and Washington, with its Utility Score being up to 50\% higher than other methods. CoS's zero-shot learning capability is benefited from its semantic understanding for scheduling events, via our knowledge distillation of the high-quality schedules. Although we only train on the New York dataset, the model indeed learns the universal and transferable spatiotemporal scheduling semantic knowledge, allowing it to interpret and manage the data from unseen cities. It is worth noting that the New York dataset actually has significantly fewer events than  Washington and London's, which further demonstrates the effective learning paradigm of CoS.

   

\begin{table}[tb]
    \centering
    \small
    \setlength{\tabcolsep}{2.2mm}
    \begin{tabular}{l|ccc}
    \toprule
        \textbf{Methods} 
        & \shortstack{\textbf{Utility}\\\textbf{Score}} 
        & \shortstack{\textbf{Latency}\\\textbf{(seconds)}} 
        & \shortstack{\textbf{Conflict}\\\textbf{Rate (\%)}} \\
    \midrule
    Default & 3.37 & 1.29 & -- \\
    w/o \textit{Exploration} & 3.15 & 0.51 & 58 \\
    w/o \textit{Verification} & 3.31 & 1.04 & 19 \\
    w/o \textit{Integration} & 3.26 & 1.13 & 24 \\
    w/o \textit{Post-processing} & 3.32 & 1.31 & 15 \\
    \bottomrule
    \end{tabular}
    \caption{Ablation study comparing our methods with baselines on the New York dataset. The conflict rate refers to the model's scheduling results prior to the removal of conflicts}
    \label{tab:ablation}
\end{table}
\subsection{Ablation Study (RQ3)}

In this section, we examine the role of each step in CoS in unleashing the potential of LLMs. Specifically, as shown in Table~\ref{tab:ablation}, we respectively ablate the three steps of CoS, \textit{i.e.}, \textit{Exploration}, \textit{Verification}, and \textit{Integration}, to observe their impacts.
W/o \textit{Exploration} means expecting the model  to directly generate the optimal schedule without comparing to other candidates. W/o \textit{Verification} means removing the verification process. W/o \textit{Integration} means randomly selecting a schedule from the candidate schedules, and w/o \textit{Post-processing} means directly removing conflicting events.

We find that removing any component significantly degrades the performance of CoS, which fully demonstrates the effectiveness of each part. Among them, \textit{Exploration} is the most important (with a 7\% drop), indicating that the initial construction of the solution space is crucial for ultimately deriving the optimal event schedule.

\subsection{Hyperparameter Sensitivity Analysis (RQ4)}

\begin{figure}[t]
    \centering 
    \includegraphics[width=1\columnwidth]{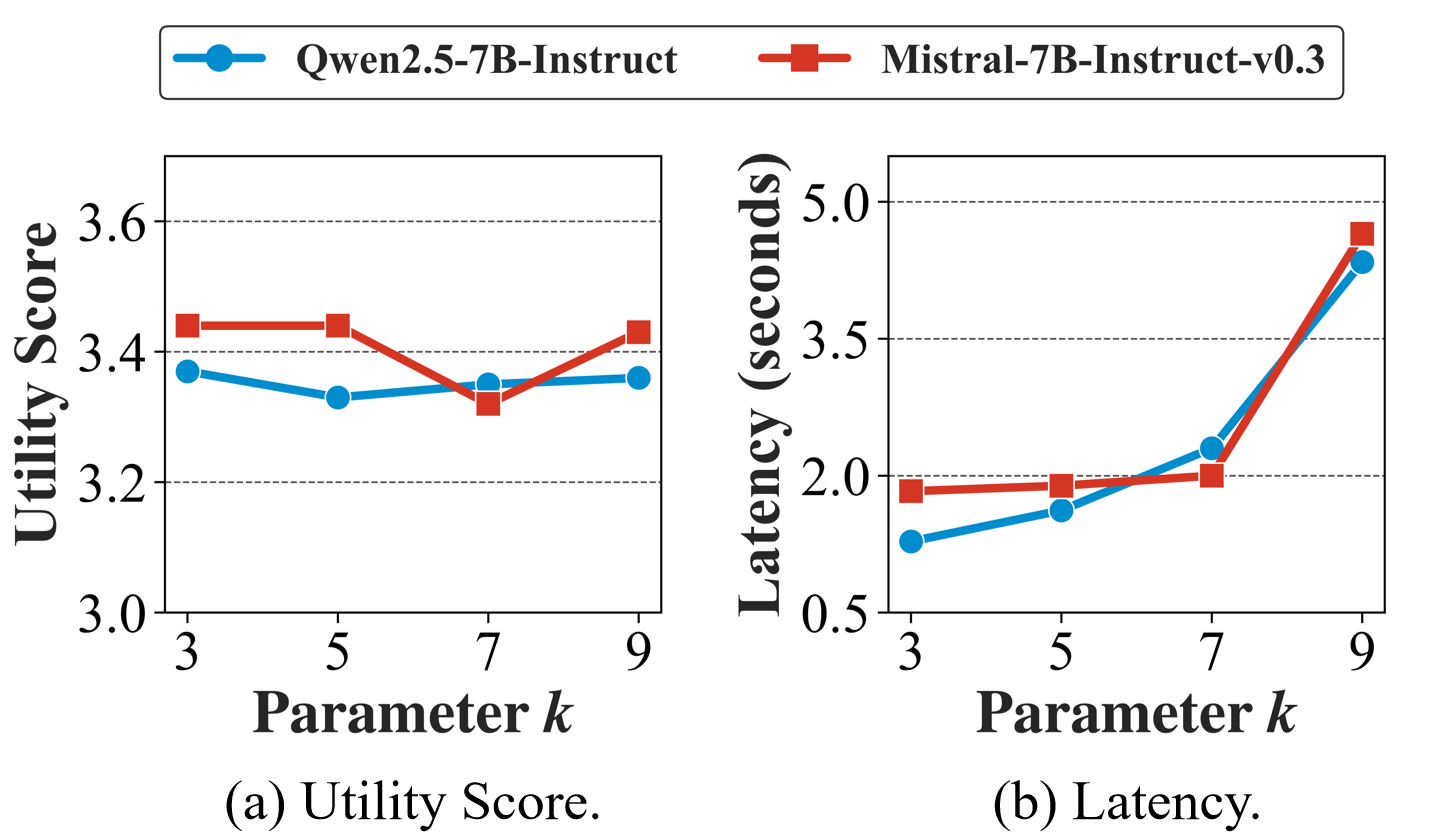} 
    \caption{Hyperparameter Sensitivity Analysis.} \label{fig:hyper_parameter_analysis} 
\end{figure}

In the CoS framework, there is a tunable parameter $k$, the number of candidate solutions in the \textit{exploration} phase. As shown in Figure~\ref{fig:hyper_parameter_analysis}, we conduct a hyperparameter sensitivity analysis on two different backbones, \textit{i.e.}, Qwen2.5-7B and Mistral-7B. The results indicate that the average utility scores remain stable at a high level across different $k$ values, while the inference time increases over a larger $k$, which is expected given the enlarged search space. Overall, we choose $k=3$, which already offers enough demonstration knowledge to the model on what is a high-quality schedule, while retaining a low latency.

\section{Conclusion}
This paper introduces CoS, which aims to stimulate the planning capabilities of LLMs. It fully stimulates the semantic understanding ability of large models for event scheduling. We organize CoS into text that LLMs can understand through text alignment, and distill the high-quality scheduling capabilities of search algorithms into LLMs via SFT, thereby efffciently and effectively completing event
scheduling tasks.
\section*{Acknowledgments}
This work is supported by the National Natural Science Foundation of China (Nos. 62472455, 62506075, U22B2060, 62276279); the Key-Area Research and Development Program of Guangdong Province (No. 2024B0101050005); the Research Foundation of the Science and Technology Plan Project of Guangzhou City (Nos. 2023B01J0001, 2024B01W0004); Guangdong Basic and Applied Basic Research Foundation (No. 2024B1515020032).
\bibliography{aaai2026}
\newpage
\section*{Appendix}

\subsection{A Bert-Alignment Analysis}
BERT is a pre-trained language model that can encode text into high-dimensional vectors to capture its semantic information. By comparing the BERT embedding vectors of two texts, their similarity can be calculated. Therefore, we align user preferences for events by calculating the similarity between user interests and event details, converting this alignment into a concrete numerical utility score. A higher utility score indicates a stronger user preference for that event.
\subsection{A.1 Utility Distribution}
\begin{figure}[thb] 
    \centering 
    \includegraphics[width=1\columnwidth]{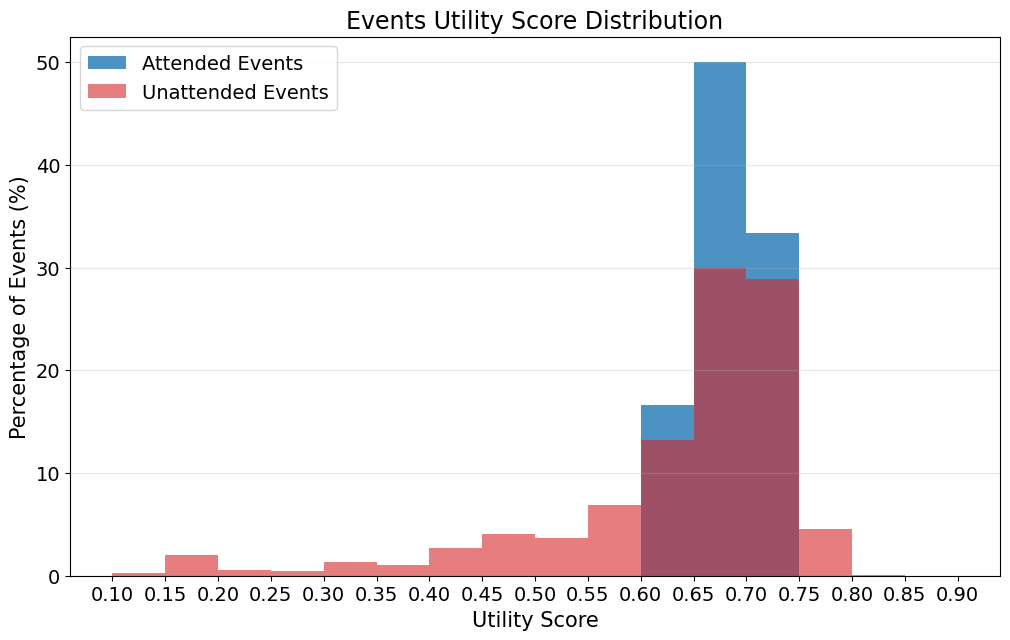} 
    \caption{\textbf{Events Utility Score Distribution.} We investigate the distribution of utility scores for events users actually attended and for randomly sampled unattended events. The utility scores for attended events are concentrated in a higher range (0.60-0.75), indicating that BERT effectively captures user preferences.} 
    \label{fig:Utility Distribution} 
\end{figure}
Figure ~\ref{fig:Utility Distribution} shows the distribution of utility scores for events that users have actually attended and for randomly sampled unattended events. We can observe that the utility scores for events actually attended by users are concentrated between 0.60 and 0.75, which is overall significantly higher than those for events users have not attended. This indicates that our BERT preference alignment can effectively capture users' actual preferences. However, there is also a very small portion of unattended events whose utility scores fall between 0.75 and 0.8. This situation might arise because the start and end times of these events conflicted with the user's other plans, or due to other personal reasons, the user was unable to actually attend these events.
\subsection{A.2 Case Study}
\begin{figure}[thb] 
    \centering 
    \includegraphics[width=1\columnwidth]{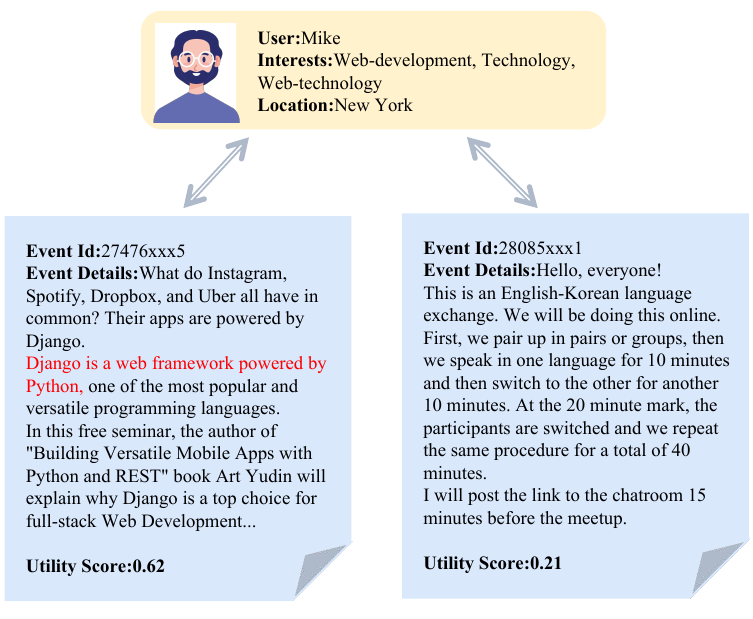} 
    \caption{\textbf{Case Study.} For a user like Mark, whose interests are in Web, a Web Programming Course receives a high utility score, whereas an English-Korean Language Exchange yields a low score.} 
    \label{fig:case_study} 
\end{figure}
We conduct a case study to specifically analyze whether BERT can align with user preferences in real-world scenarios, as shown in Figure ~\ref{fig:case_study}. In this example, the actual user, Mark (pseudonym), has interests in Web-development, Technology, and Web-technology. When event information closely matches these interests, such as Web Programming Course, the utility score aligned by BERT is high. Conversely, for events with disparate topics, such as English-Korean Language Exchange, the utility score aligned by BERT is significantly lower.
\subsection{A.3 Event Participation Utility Comparison}
\begin{figure}[thb] 
    \centering 
    \includegraphics[width=1\columnwidth]{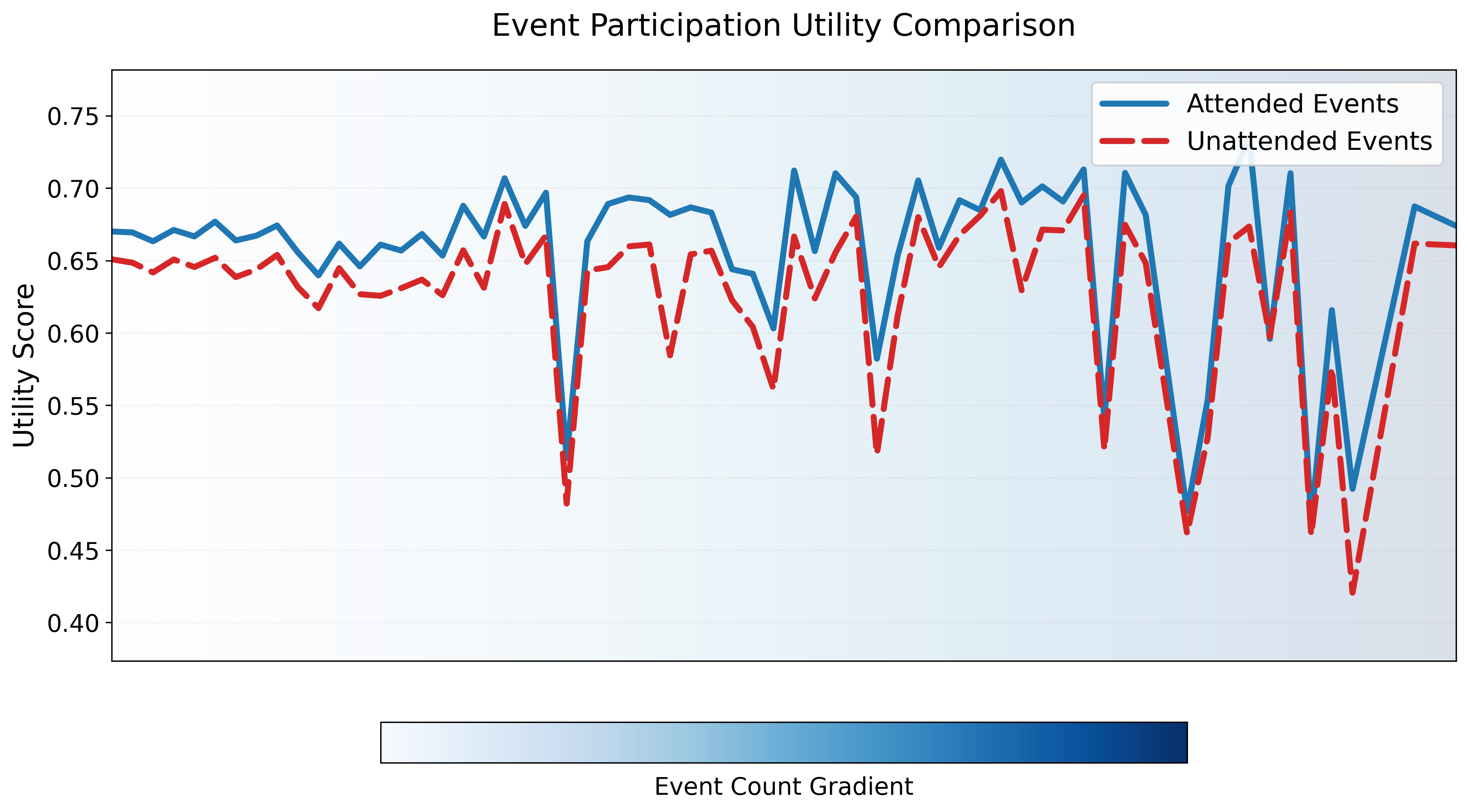} 
    \caption{\textbf{User Engagement and Event Preference Alignment Analysis.}This figure illustrates that as the number of events a user actually participates in increases, the average utility score for their attended events consistently remains higher than that for unattended events. This indicates that the BERT-driven preference alignment mechanism effectively identifies and recommends highly matched events for users, regardless of their activity level.} 
    \label{fig:line} 
\end{figure}
To further analyze the performance of the BERT-driven preference alignment mechanism under varying user activity levels, we investigate the impact of the number of events a user has attended on their average utility score. As Figure ~\ref{fig:line} illustrates, the horizontal axis of the graph represents the number of events a user actually participates in, with the color gradually transitioning from light to dark to indicate an increasing number of attended events. The vertical axis represents the average utility score. Two lines are plotted in the figure: a solid line representing the average utility score for events the user attends, and a dashed line representing the average utility score for events the user does not attend (calculated by randomly sampling a certain number of unattended events for each user). As the figure shows, we find that as the number of events a user attends increases, the average utility score for attended events consistently remains higher than that for unattended events. This indicates that BERT preference alignment can plan highly matched events for users across different activity levels.
\section{B NP-Hard Proof for Event Scheduling.}

\paragraph{Preliminaries.}
\textbf{Event Scheduling.} Given an event set $E$ and a user $u_j$, the recommendation task on EBSN is to generate an event sequence $T = \langle e_{i_1} \rightarrow e_{i_2} \rightarrow \cdots \rangle$ that maximizes the total utility score while retaining validity:
\begin{gather}
    \arg\max_{T} \sum_{e_i \in T} s_{i,j} \nonumber. 
\end{gather}
\textbf{Feasible movement.}
\begin{equation} \label{eq:feasible_movement}
    \forall (e_{i} \rightarrow e_{i'}) \in T,\quad t_{i'}^{\text{start}} - t_{i}^{\text{end}} \geq t(\text{loc}_{i}, \text{loc}_{i'}),
\end{equation}
where the function $t(\cdot)$ denotes the traveling time between the two locations.
\\\textbf{No time conflicts.}
\begin{equation} \label{eq:no_conflict}
    \forall e_i, e_j \in T \ (i \neq j), \quad
    [t_i^{\text{start}}, t_i^{\text{end}}] \cap [t_j^{\text{start}}, t_j^{\text{end}}] = \emptyset.
\end{equation}

\paragraph{Proof.}
We reduce the \textbf{Directed Hamiltonian Path (DHP)} problem to \textbf{Event Scheduling Problem (ESP)}.

\begin{enumerate}
    \item \textbf{Reduction Construction.} 
    \\Given a DHP instance $G = (V, E)$ with $|V| = n$, we construct an ESP instance as follows:
    \begin{itemize}
        \item For each vertex $v_i \in V$, create an event $e_i$ with:
        \begin{itemize}
            \item Time window $[i, i+1)$,
            \item Utility $s_i = 1$,
            \item Travel time:
            \[
            t(\text{loc}_i, \text{loc}_j) =
            \begin{cases}
            1 & \text{if } (v_i, v_j) \in E, \\
            +\infty & \text{otherwise}.
            \end{cases}
            \]
        \end{itemize}
    \end{itemize}

    \item \textbf{Solution Equivalence.}
    \begin{itemize}
        \item \textbf{DHP $\Rightarrow$ ESP}. A Hamiltonian path $\langle v_{k_1}, \dots, v_{k_n} \rangle$ in $G$ maps to a valid schedule:
        \begin{align*}
            \textbf{No conflicts}: &\ [t_{k_i}^{\text{start}}, t_{k_i}^{\text{end}}) \cap [t_{k_j}^{\text{start}}, t_{k_j}^{\text{end}}) = \emptyset \quad \forall i \neq j, \\
            \textbf{Feasible movement}: &\ t_{k_{i+1}}^{\text{start}} - t_{k_i}^{\text{end}} = 1 \ge t(\text{loc}_{k_i}, \text{loc}_{k_{i+1}}).
        \end{align*}

        \item \textbf{ESP $\Rightarrow$ DHP}: A schedule with utility $n$ must include all events in an order $\langle e_{k_1}, \dots, e_{k_n} \rangle$ where
        \[
        t(\text{loc}_{k_i}, \text{loc}_{k_{i+1}}) = 1 \implies (v_{k_i}, v_{k_{i+1}}) \in E.
        \]
    \end{itemize}
\end{enumerate}

\noindent\textbf{Conclusion.}
\begin{itemize}
    \item The reduction is polynomial-time ($O(n^2)$ to assign $t(\cdot)$).
    \item DHP is NP-Hard $\Rightarrow$ ESP is NP-Hard.
\end{itemize}

\section{C Experimental supplement}
\subsection{C.1 Prompts}
The COT prompts and non-COT prompts (referred to as direct) are as follows. In both cases, event\_list represents the input event information. Given the inherent challenges for oversized LLMs in processing geographical constraints, our prompts are designed to exclusively task these models with handling temporal constraints. Subsequent processing of both geographical and temporal conflicts will be performed by our dedicated post-processing method.
\begin{figure}[thb] 
    \centering 
    \includegraphics[width=1\columnwidth]{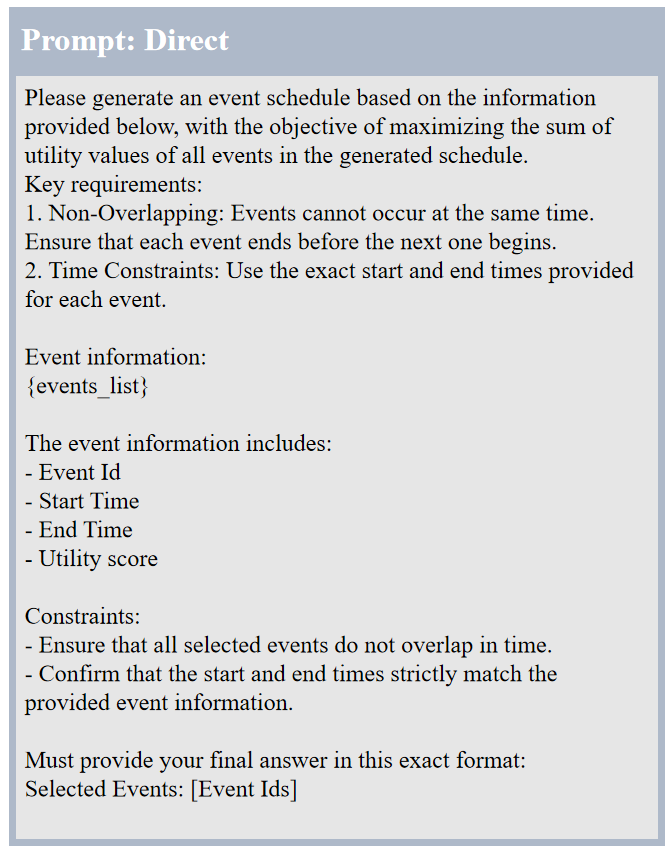} 
    \label{fig:prom2} 
\end{figure}
\begin{figure}[thb] 
    \centering 
    \includegraphics[width=1\columnwidth]{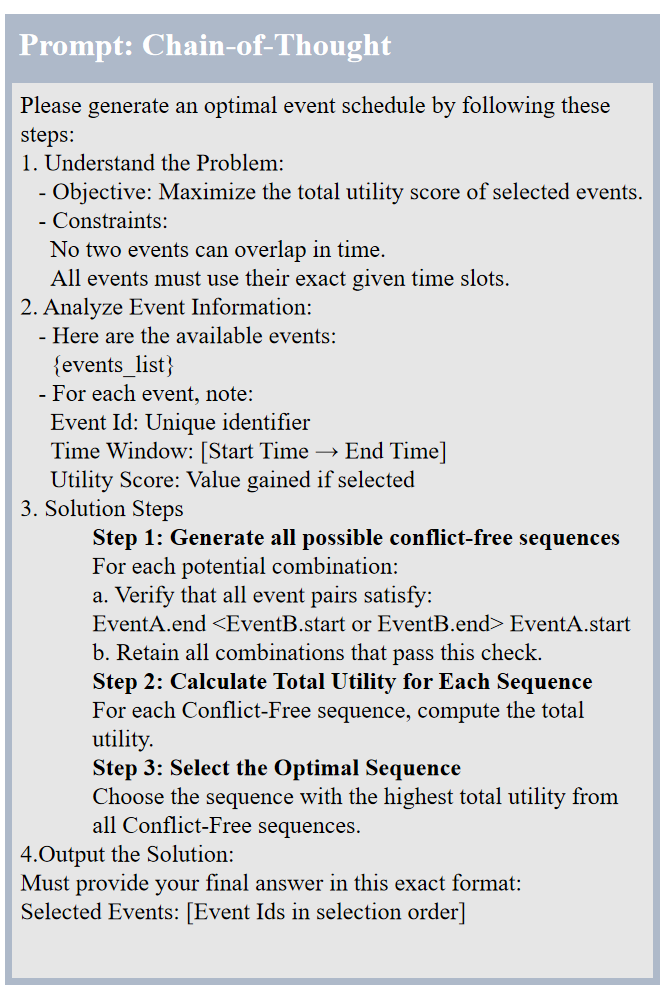} 
    \label{fig:prom1} 
\end{figure}

\subsection{C.2 Answer Comparison}

We select a representative algorithm case from each method category for comparison, the event information input for these methods is identical. Analysis reveals that the greedy algorithm, by considering only the locally optimal solution at each step (i.e., maximizing the current utility score), cannot achieve global optimality. The GNN-DRL method, limited by its model parameter scale, exhibits general performance when handling complex problems. Qwen-Max, combined with Chain-of-Thought, has shortcomings in processing numerical information, leading to numerous conflicts regarding time constraints. In contrast, our CoS (Chain-of-Scheduling) demonstrates superior experimental results.
\begin{figure}[thb]
    \centering 
    \includegraphics[width=1\columnwidth]{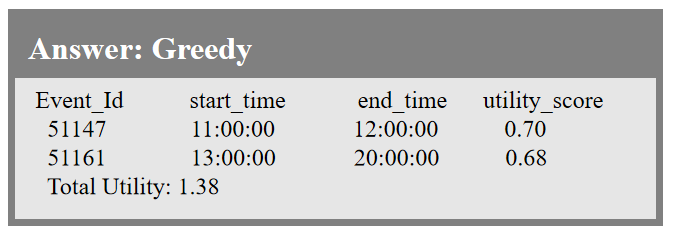} 
    \label{fig:greedy} 
\end{figure}
\begin{figure}[thb] 
    \centering 
    \includegraphics[width=1\columnwidth]{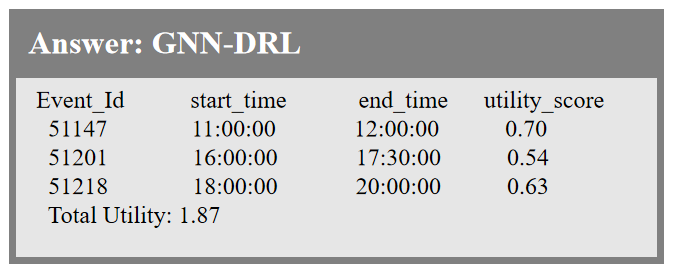} 
    \label{fig:GNN-RL} 
\end{figure}
\begin{figure}[thb]
    \centering
    \begin{subfigure}[b]{1\columnwidth} 
        \centering
        \includegraphics[width=\textwidth]{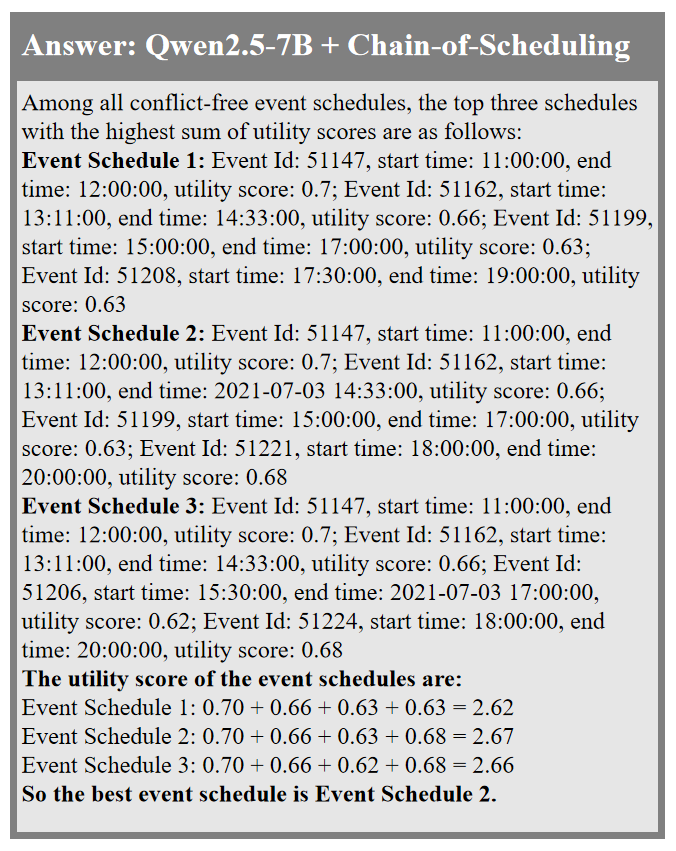}
        \label{fig:cos}
    \end{subfigure}

    \begin{subfigure}[b]{1\columnwidth} 
        \centering
        \includegraphics[width=\textwidth]{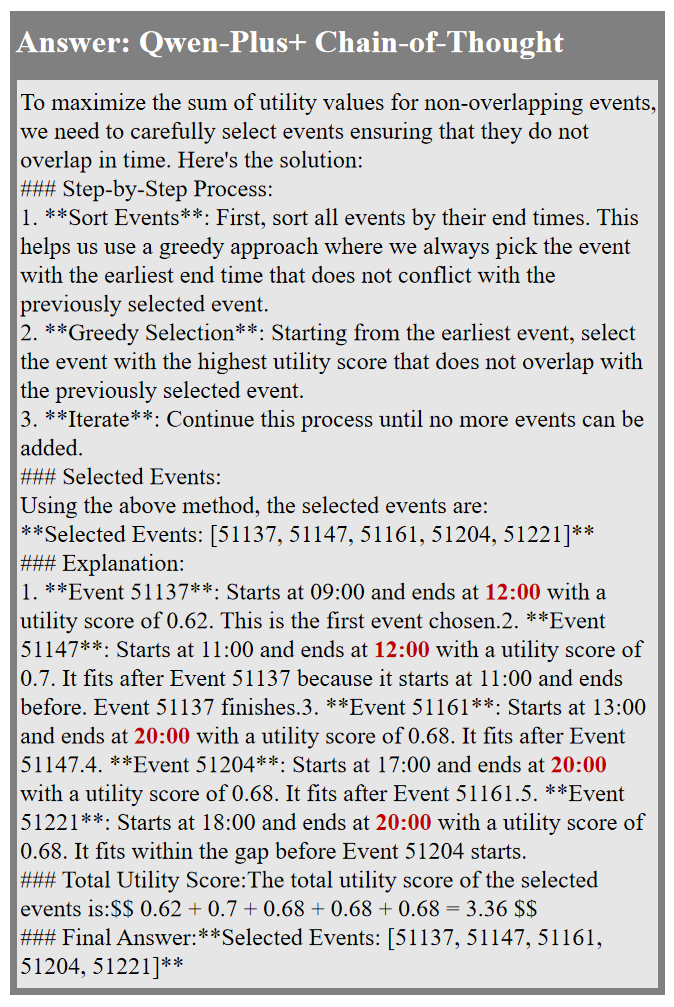}
        \label{fig:COT}
    \end{subfigure}
    \label{fig:combined_figures} 
\end{figure}

\end{document}